\documentclass{article}
\usepackage{enumitem}

\usepackage[table]{xcolor}

\usepackage{times}
\usepackage{soul}
\usepackage{url}
\usepackage{makecell}
\usepackage{hyperref}
\usepackage{graphicx}
\usepackage[small]{caption}
\usepackage{amsmath}
\usepackage{amsthm}
\usepackage[switch]{lineno}
\usepackage{longtable}
\usepackage{supertabular}
\usepackage{multirow}
\usepackage{tikz}                  
\usepackage[edges]{forest}

\usepackage[in]{fullpage}  
\usepackage{arxiv} 
\usepackage{mathpazo} 

\usepackage[utf8]{inputenc} 
\usepackage[T1]{fontenc}    
\usepackage{url}            
\usepackage{booktabs}       
\usepackage{amsfonts}       
\usepackage{microtype}      
\usepackage{lipsum}         

\usepackage{graphicx}  
\usepackage{float} 
\usepackage{subfigure} 
\usepackage{algorithm}
\usepackage{algorithmic}
\usepackage{tabulary}

\usepackage{amssymb}
\makeatletter

\newcommand{\Rmnum}[1]{\expandafter\@slowromancap\romannumeral #1@}
\makeatother

\usepackage{pifont}

\hypersetup{
    colorlinks=true,
    linkcolor=green,
    citecolor=red
}
\usepackage{tcolorbox}

\usepackage{natbib}
\usepackage{bbding}

\tcbset{
  mydiscussbox/.style={
    colback=gray!10!white,
    colframe=gray!80!black,
    coltitle=black,
    fonttitle=\bfseries,
    title=Discussion,
    boxrule=0.8pt,
    arc=4pt,
    left=6pt,
    right=6pt,
    top=6pt,
    bottom=6pt,
    enhanced,
    sharp corners=south
  }
}

\newtcolorbox{discussionbox}[1][]{%
  colback=gray!5!white,
  colframe=blue!60!black,
  fonttitle=\bfseries,
  title=Discussion,
  top=8pt,
  bottom=8pt,
  left=8pt,
  right=8pt,
  #1
}

\hypersetup{
    colorlinks=true,
    citecolor=blue,
    linkcolor=blue,
    citecolor=orange
}

\definecolor{red}{rgb}{1, 0, 0}

\definecolor{green}{rgb}{0, 1, 0}

\definecolor{blue}{rgb}{0, 0, 1}

\newcommand{\Tag}[2]{%
    \tikz[baseline=(X.base)]{
        \node[draw, rounded corners, fill=#1!30, inner sep=2pt] (X) {\textbf{#2}};
    }
}

\title{A Survey on Large Language Models for Mathematical Reasoning}

\usepackage{authblk}
\author{%
  Peng-Yuan Wang\textsuperscript{\rm 1,2*}, Tian-Shuo Liu\textsuperscript{\rm 1,2*}, 
  Chenyang Wang\textsuperscript{\rm 1,2},
Yi-Di Wang\textsuperscript{\rm 1,2}, Shu Yan\textsuperscript{\rm 1}, Cheng-Xing~Jia\textsuperscript{\rm 1,2},  Xu-Hui Liu\textsuperscript{\rm 1,2}, Xin-Wei Chen\textsuperscript{\rm 3}, Jia-Cheng~Xu\textsuperscript{\rm 4,5}, 
Ziniu Li\textsuperscript{\rm 6}, Yang Yu\textsuperscript{\rm 1,2,$\diamond$}\\
  \textsuperscript{\rm 1} National Key Laboratory for Novel Software Technology, Nanjing University, China \\
  \textsuperscript{\rm 2} School of Artificial Intelligence, Nanjing University, China \\
  \textsuperscript{\rm 3} Polixir.ai \\
    \textsuperscript{\rm 4} Nanyang Technological University, Singapore \\
    \textsuperscript{\rm 5} Skywork AI, Singapore \\
    \textsuperscript{\rm 6} The Chinese University of Hong Kong, Shenzhen \\
  \textsuperscript{*} Equal contribution\\
  \textsuperscript{$\diamond$} Corresponding: yuy@nju.edu.cn
}

\date{}
\begin{document}

\maketitle

\begin{abstract}

Mathematical reasoning has long represented one of the most fundamental and challenging frontiers in artificial intelligence research. In recent years, large language models (LLMs) have achieved significant advances in this area. This survey examines the development of mathematical reasoning abilities in LLMs through two high-level cognitive phases: comprehension, where models gain mathematical understanding via diverse pretraining strategies, and answer generation, which has progressed from direct prediction to step-by-step Chain-of-Thought (CoT) reasoning. We review methods for enhancing mathematical reasoning, ranging from training-free prompting to fine-tuning approaches such as supervised fine-tuning and reinforcement learning, and discuss recent work on extended CoT and ``test-time scaling''. Despite notable progress, fundamental challenges remain in terms of capacity, efficiency, and generalization. To address these issues, we highlight promising research directions, including advanced pretraining and knowledge augmentation techniques, formal reasoning frameworks, and meta-generalization through principled learning paradigms. This survey tries to provide some insights for researchers interested in enhancing reasoning capabilities of LLMs and for those seeking to apply these techniques to other domains.
\end{abstract}

\section{Introduction}
 ``Can machines think?'' This profound question, posed by Alan Turing in the 1950s~\citep{turing1950computing}, established the philosophical foundation for modern artificial intelligence. Since then, enabling machines to reason has remained a central and enduring objective in AI research. Among the most rigorous and illuminating assessments of machine reasoning is mathematical problem-solving, a domain that demands not only the manipulation of symbols, but also the representation and understanding of abstract concepts, the construction of formal arguments, and the transfer of principles to new and varied contexts. As such, mathematical reasoning provides a precise and demanding lens through which to evaluate and advance machine intelligence. Beginning in the 1960s, AI researchers sought to endow machines with mathematical reasoning abilities by developing systems capable of representing and manipulating formal knowledge. Early work centered on symbolic rule-based systems~\citep{feigenbaum1963computers, bobrow1964natural}, which depended on handcrafted rules and pattern matching~\citep{slagle1965experiments, fletcher1985understanding}. While pioneering, these approaches were restricted to narrow domains and lacked generalization. Subsequent efforts, such as semantic parsing methods~\citep{kwiatkowski2013scaling, goldwasser2014learning}, focused on mapping problem text to structured logical forms, but continued to rely heavily on human engineering and struggled to scale to the full diversity of mathematical tasks.

\begin{figure}
    \centering
    \includegraphics[width=0.9\linewidth]{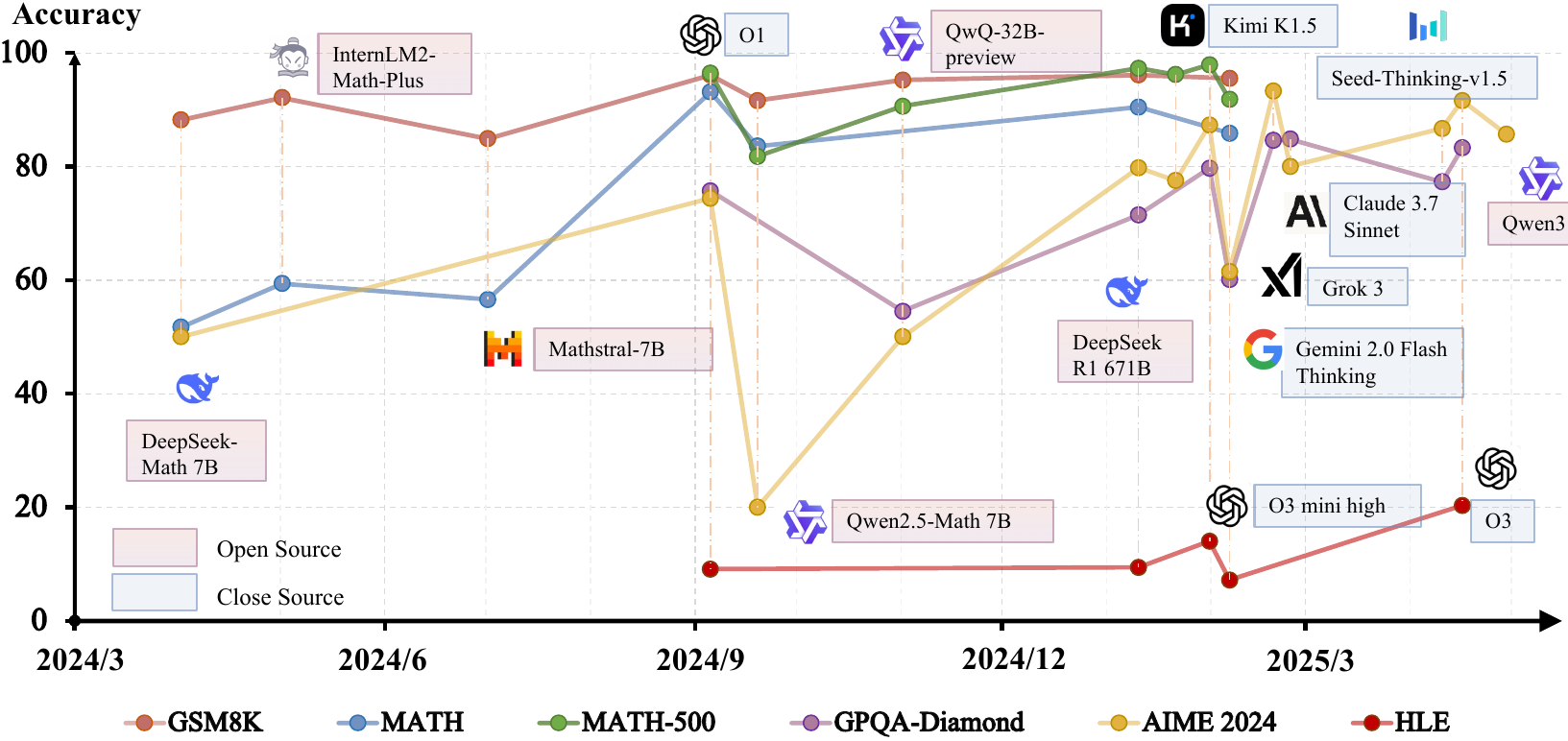}
    \caption{The rapid advancements in mathematical capabilities in recent years.}
    \label{fig:progress}
\end{figure}

Recent advances in large language models (LLMs) have profoundly reshaped the field of natural language representation and understanding. The introduction of transformer-based architectures and instruction-tuned models such as ChatGPT has led to remarkable progress in natural language problem-solving. In this context, mathematical reasoning, a long-standing and rigorous benchmark for AI systems, has become a focal point for evaluating LLM capabilities. As illustrated in Figure~\ref{fig:progress}, state-of-the-art models now exhibit significantly enhanced performance across a wide range of complex mathematical benchmarks.
 For example, Grok 3 Beta achieved an impressive score of $83.9\%$ on the AIME 2024~\citep{aime2024}, a prestigious Olympiad-level contest. This result places the model within the top $2.5\%$ of all participants nationwide\footnote{Human performance is based on data from MAA: \url{https://maa.edvistas.com/eduview}.}, as shown in Figure~\ref{fig:performance_modelsize}. As LLMs continue to excel in mathematical problem-solving, their impact is becoming increasingly significant across various domains. In Finance, they can leverage complex mathematical reasoning capabilities to process both structured tables and unstructured text, enabling Financial Document Question Answering, leading to handling complex mathematical scenarios~\citep{DBLP:conf/acl/SrivastavaM0GR24}.  In the Medical field, advanced LLMs can work as a medical diagnostic assistant, achieving high accuracy in answering specialized healthcare-related questions~\citep{10.1007/978-981-97-1717-0_12}. Driven by the rapid advancements and increasing applicability of mathematical reasoning in LLMs, this work seeks to consolidate the fragmented developments into a coherent framework.

The ability of LLMs to perform mathematical reasoning can be grounded roughly in two elements: comprehension of mathematical concepts and the generation of solutions through step-by-step deduction. Comprehension involves acquiring a broad and flexible understanding of diverse mathematical concepts, problem formats, and difficulty levels to enable effective reasoning and problem-solving. Recent research has demonstrated that large-scale pre-training on mathematical corpora, such as textbooks, academic publications, and problem datasets, enables LLMs to internalize domain-specific knowledge, terminology, and the contextual reasoning patterns characteristic of mathematical discourse~\citep{shao2024deepseekmath, yang2024qwen2}. This data-driven paradigm marks a significant shift from earlier approaches based on rule-matching~\citep{slagle1965experiments, fletcher1985understanding} or semantic parsing~\citep{kwiatkowski2013scaling, goldwasser2014learning}, moving toward a more holistic and contextualized understanding of mathematics.
The second element, solution generation, focuses on producing logically coherent intermediate steps during problem-solving. A key technique for achieving this in LLMs is chain-of-thought (CoT) prompting~\citep{cotreasoning}, which encourages models to emulate deductive reasoning by generating step-by-step explanations. This approach has proven effective in enabling LLMs to tackle more complex, abstract, and real-world mathematical challenges, as it structures the reasoning process into interpretable and verifiable steps.

Despite their impressive language understanding abilities, pre-trained LLMs often struggle to produce contextually appropriate and relevant responses when applied directly to downstream tasks, frequently resulting in repetitive or irrelevant output. To mitigate these shortcomings, prompt engineering, particularly approaches like In-Context Learning (ICL)~\citep{jie2023leveraging, teaching}, has emerged as a vital strategy for enhancing the reasoning capabilities of LLMs through careful design of input prompts.
In addition to prompt engineering, fine-tuning\footnote{Here, we refer supervised fine-tuning, reinforcement learning, and any other stages of guided learning after pre-training of LLMs as ``fine-tuning'' instead of ``post-training'', for exactness of verbal expression.} strategies have been shown to further boost model performance. For example, supervised fine-tuning (SFT) on high-quality demonstrations adapts pre-trained models to specific tasks or domains~\citep{yang2024qwen2, ho2023large, magister2023teaching}. This process helps align models with instruction-following objectives, although it introduces challenges such as overfitting and limitations on exploration~\citep{li2025preserving, wang2planning, zeng2025simplerl}. Reinforcement learning (RL) methods~\citep{guo2025deepseek} further empower models to improve their problem-solving abilities through trial-and-error exploration.
Recent research has also shown that extending CoT reasoning by generating longer token sequences during inference~\citep{openai2024o1, snell2024scaling} can significantly enhance the reasoning capabilities of LLMs, resulting in more structured and accurate solutions. Techniques such as search-based methods~\citep{tot, twistmcts} and reinforcement fine-tuning~\citep{openai2024o1, luong2024reft} can be incorporated to further refine the model’s reasoning abilities, especially when working with extended CoT reasoning.

Recently, a wide range of methods and models have been developed to advance mathematical reasoning. Several surveys have reviewed aspects of this field. For example, \citet{zhang2019gap} and \citet{meadows2022survey} focus on traditional approaches to mathematical problem-solving. Other works, such as~\cite{pro_chan_survey} and~\cite{lu2023survey}, primarily discuss datasets and fine-tuning methods. The survey by~\cite{yan2024survey} summarizes progress in multi-modal mathematical reasoning based on LLMs. In the broader context of reasoning, \citet{li2025system} offers a systematic analysis of advancements in System 2 thinking, while~\citet{xu2025towards} introduces the application of reinforcement learning to reasoning tasks. Additionally, \citet{zhou2025reinforced} reviews RL-based multi-modal reasoning. This survey tries to complement these works by offering a comprehensive and up-to-date overview of recent developments in LLM-based mathematical reasoning, helping to synthesize insights across this rapidly evolving field.

\tikzstyle{subsubsection}=[font=\small, align=center, text width=13em]

\begin{figure*}[t]
    \centering
    \resizebox{0.85\textwidth}{!}{
        \begin{forest}
            forked edges,
            for tree={
                grow=east,
                reversed=true,
                anchor=base west,
                parent anchor=east,
                child anchor=west,
                base=center,
                font=\large,
                rectangle,
                draw=black,
                rounded corners,
                align=left,
                minimum width=4em,
                edge+={black, line width=1pt, rounded corners=5pt},
                l sep=24.5pt,
                s sep=3pt,
                inner xsep=2pt,
                inner ysep=3pt,
                line width=0.8pt,
                ver/.style={rotate=90, child anchor=north, parent anchor=south, anchor=center},
            },
            where level=1{text width=12em,font=\small,}{},
            where level=2{text width=12em,font=\small,}{},
            [
              \textbf{Mathematical Reasoning with LLM}, font=\small, align=center
                [
                \textbf{Background} ($\S$\ref{sec:background}), draw=blue,fill=cyan!10, text width=20em
                [
                  \textbf{Pre-training} ($\S$\ref{subsec:pretrain}), draw=orange,fill=yellow!10, text width=24.5em
                ]
                [
                  \textbf{Fine-tuning: Supervised Fine-tuning} ($\S$\ref{subsec:fine-tuning-sft}), draw=orange,fill=yellow!10, text width=24.5em
                ]
                [
                  \textbf{Fine-tuning: Reinforcement Learning} ($\S$\ref{subsec:finetuning-rl}), draw=orange,fill=yellow!10, text width=24.5em
                ]
                [
                  \textbf{Prompting LLMs for Reasoning} ($\S$\ref{subsec:prompting}), draw=orange,fill=yellow!10, text width=24.5em
                ]
               ]
              [
                \textbf{Comprehension and Generation} ($\S$\ref{sec:comprehension}), draw=blue,fill=cyan!10, text width=20em
                [
                  \textbf{Diverse Mathematical Representation Formats} ($\S$\ref{sec:Mathematical Representation Formats}), draw=orange,fill=yellow!10, text width=24.5em
                ]
                [
                  \textbf{Chain-of-Thought: A Key Driver of LLM Generation} ($\S$\ref{subsubsec:prompting-cot}), draw=orange,fill=yellow!10, text width=24.5em
                ]
               ]
               [
                    \textbf{Methods for Boosting Reasoning} ($\S$\ref{sec:boosting}), draw=blue,fill=cyan!10, text width=20em
                    [
                        \textbf{Supervised Fine-Tuning} ($\S$\ref{subsec:SFT}), draw=orange,fill=yellow!10, text width=24.5em
                    ]
                    [ 
                      \textbf{Reinforcement Learning} ($\S$\ref{subsec:RL}), draw=orange,fill=yellow!10, text width=24.5em
                    ]
                    [ 
                      \textbf{Test Time Inference with Structural Search} ($\S$\ref{subsec:testing-search}), draw=orange,fill=yellow!10, text width=24.5em
                    ]
                    [ 
                      \textbf{Self-Improvement} ($\S$\ref{subsubsec:selfimprove}), draw=orange,fill=yellow!10, text width=24.5em
                    ]
                    [ 
                      \textbf{Utilizing External Knowledge} ($\S$\ref{subsec:RAG}), draw=orange,fill=yellow!10, text width=24.5em
                    ]
                ]
                [
                   \textbf{Discussions} ($\S$\ref{sec:discussion}), draw=blue,fill=cyan!10, text width=20em
                ]
                [
                   \textbf{Potential Directions} ($\S$\ref{sec:challenge}), draw=blue,fill=cyan!10, text width=20em
                ]
            ] 
        \end{forest}
    }
    \caption{The organization of this survey.}
    \label{fig:overview}
\end{figure*}
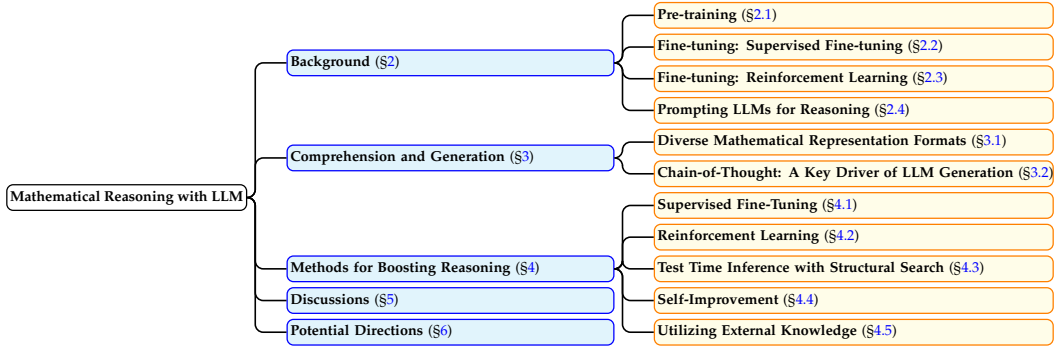

In this paper, we review the research in the field of mathematical reasoning as illustrated in Figure~\ref{fig:overview}. First, in Section \ref{sec:background}, we give a brief introduction to the background of LLM. In Section \ref{sec:comprehension}, we summarize recent innovations and identify some essential elements that enable mathematical reasoning. 
Section \ref{sec:boosting} reviews related methods, which can boost reasoning in LLM, including prompting, fine-tuning, test-time scaling, and self-improvement. Finally, Sections \ref{sec:discussion} and \ref{sec:challenge} discuss current limitations and potential directions for future research.

\section{Background}
\label{sec:background}
\subsection{Pre-Training}
\label{subsec:pretrain}
Large language models (LLMs) acquire knowledge and develop general language understanding across various domains through pre-training on extensive corpora~\citep{zhao2023survey}. Given a dataset $\mathcal{D}_{pre}=\{x^{(i)}\}_{i=1, \ldots, N}$, where $x^{(i)}$ represents the $i$-th sequence of tokens in the dataset, the pre-training optimization objective is typically a form of language modeling, such as predicting the next token:
\begin{equation}
\max _{\theta} \sum_{x \in \mathcal{D}_{pre}} \sum_{t=1}^{|x|} \log \left(\pi_{\theta}\left(x_t \mid x_{<t}\right)\right)
\end{equation}
where $\pi_{\theta}$ is the model parameterized by $\theta$, $x_t$ is the token in sequence $x$ at position $t$, and $x_{<t}$ are the preceding tokens.

During the pre-training phase, access to high-quality and highly diverse data is fundamental in shaping the reasoning, comprehension, and generalization abilities of LLMs~\citep{yang2024qwen, touvron2023llama}. Vast amounts of text corpora, including books, papers, websites, and code, can be leveraged to improve the performance of LLMs. However, raw data often contains noise and inconsistencies, making effective data selection and filtering techniques~\citep{shao2024deepseekmath, joulin2016fasttext,zhang2024autonomous,lin2024rho} important for curating high-quality information. Beyond relying solely on raw data, recent developments in LLM training have incorporated the use of synthetic data~\citep{wang2023learning,ying2024internlm, zhou2024jiuzhang3, yang2024qwen2}. By combining high-quality raw data with synthetically generated examples, LLMs can potentially achieve enhanced capabilities in areas such as reasoning.

\subsection{Fine-Tuning: Supervised Fine-Tuning}
\label{subsec:fine-tuning-sft}
To adapt LLMs to specific downstream tasks or to align their behavior with human instructions, a common stage is supervised fine-tuning (SFT). Given a dataset $\mathcal{D}_{sft}=\{(x^{(i)}, y^{(i)})\}_{i=1, \ldots, M}$, where $x_i$ is an instructional prompt or input sequence and $y_i$ is the desired output sequence, SFT refines the model by optimizing a similar language modeling objective:
\begin{equation}
\max _{\theta} \sum_{(x, y) \in \mathcal{D}_{sft}} \sum_{t=1}^{|y|} \log \left(\pi_{\theta}\left(y_t \mid x, y_{<t}\right)\right)
\end{equation}
This process trains the model to mimic the ``golden'' responses provided in the SFT dataset.

The choice and quality of the task-specific dataset play a significant role in the effectiveness of SFT. Methods such as data augmentation and data synthesis are commonly used to enhance dataset quality and diversity. To further enhance model performance, fine-tuning can incorporate knowledge distillation~\citep{hinton2015distilling}, where a student model learns from the outputs of a more capable teacher model. The teacher’s outputs can provide additional supervisory signals beyond ground truth labels, potentially improving fluency, consistency, and efficiency.

\subsection{Fine-Tuning: Reinforcement Learning}
\label{subsec:finetuning-rl}
While SFT helps LLMs learn to follow instructions and generate relevant text, it may not always be sufficient to align model outputs with complex human preferences, ensure factual accuracy, or reduce undesirable behaviors like generating harmful or biased content~\citep{openai2023gpt4systemcard, google2023geminitechnicalreport}. Reinforcement learning (RL) offers a framework to further refine LLM behavior based on broader notions of quality, often captured by a reward signal~\citep{ziegler2019fine}. A prominent application of RL in this domain is reinforcement learning from human feedback (RLHF), where human preferences are used to train a reward model, which then guides the LLM's fine-tuning process~\citep{christiano2017deep, bai2022training, trainlmfollowinstruct}. This approach allows the LLM to learn from scalar feedback signals that can represent nuanced aspects of response quality, such as helpfulness, harmlessness, and honesty, which can be difficult to specify directly in an SFT objective~\citep{askell2021general, menick2022teaching}.

The RL framework is typically formalized as a Markov Decision Process (MDP). An MDP is formally represented as a tuple $\mathcal{M} = (\mathcal{S},\mathcal{A},\mathcal{P},r,\gamma, \rho_0, T)$, where:
\begin{itemize}[itemsep=2pt]
    \item $\mathcal{A}$ represents the action space. For an LLM, an action $a_t$ is typically the selection of the next token from the vocabulary~\citep{trainlmfollowinstruct}.
    \item $\mathcal{S}$ represents the state space. In the context of LLMs, a state $s_t$ at time step $t$ can be defined as the sequence of tokens generated so far, often including the initial prompt: $s_t = (x, a_1, a_2, \ldots, a_{t-1})$~\citep{trainlmfollowinstruct}.
    \item $\mathcal{P}: \mathcal{S}\times \mathcal{A}\times \mathcal{S}\rightarrow [0,1]$ represents the state transition probability function, $P(s_{t+1} | s_t, a_t)$. In LLM generation, the transition is often deterministic: generating token $a_t$ in state $s_t$ leads to a unique next state $s_{t+1} = (s_t, a_t)$. Thus, $P(s_{t+1} | s_t, a_t) = 1$ if $s_{t+1}$ is the concatenation of $s_t$ and $a_t$, and $0$ otherwise.
    \item $r: \mathcal{S}\times \mathcal{A}\rightarrow \mathbb{R}$ represents the reward function. This function provides a scalar feedback signal. In RLHF, this reward is often given by a separate reward model trained on human preference data, which evaluates the quality of generated sequences~\citep{christiano2017deep, trainlmfollowinstruct}. Rewards can be sparse (e.g., given only at the end of a sequence) or dense (e.g., per token).
    \item $\gamma \in [0,1]$ is the discount factor, which balances the importance of immediate versus future rewards.
    \item $\rho_0: \mathcal{S}\rightarrow [0,1]$ represents the initial state distribution. For LLMs, this is typically determined by the distribution of input prompts $x$.
    \item $T$ denotes the horizon or maximum episode length (e.g., maximum sequence length).
\end{itemize}
The agent, in this case the LLM (also referred to as the policy $\pi_{\theta}(a_t|s_t)$), aims to learn a policy that maximizes the expected cumulative discounted reward:
\begin{equation}
    \max_{\theta} \mathbb{E}_{\tau \sim \pi_{\theta}} \left[ \sum_{t=0}^{T-1} \gamma^t r(s_t, a_t) \right]
\end{equation}
where $\tau = (s_0, a_0, s_1, a_1, \ldots, s_{T-1}, a_{T-1}, s_T)$ is a trajectory (a sequence of states and actions), $s_0 \sim \rho_0$, $a_t \sim \pi_{\theta}(\cdot|s_t)$, and $s_{t+1} \sim \mathcal{P}(\cdot|s_t, a_t)$.

The reward function $r(s_t, a_t)$ is crucial. It can be categorized into two types:
\begin{itemize}[itemsep=2pt]
    \item \textbf{Outcome-based rewards}: These are assigned based on the final result (e.g., evaluating the complete generated sequence for helpfulness or correctness)~\citep{rlhf}.
    \item \textbf{Process-based rewards}: These evaluate intermediate steps or the reasoning process leading to the outcome, which can provide more granular feedback~\citep{wang2024chain}.
\end{itemize}
Common RL algorithms like Proximal Policy Optimization (PPO) were often used to optimize the LLM policy based on the rewards from the reward model~\citep{ppo, trainlmfollowinstruct, bai2022training}. The overall goal is to steer the LLM towards generating outputs that are more aligned with desired characteristics that might be underspecified or difficult to learn through SFT alone~\citep{openai2023gpt4systemcard, google2023geminitechnicalreport}. It is worth noticing that, once the reward model is given, RLHF can train LLMs without response data, significantly improving the generalization ability of LLMs.

\subsection{Prompting LLMs for Reasoning}
\label{subsec:prompting}
Prompting has emerged as a simple yet effective way for eliciting and enhancing the reasoning capabilities of large language models (LLMs). The initial prompting paradigm, exemplified by models such as GPT-2~\citep{radford2019language}, was characterized as the zero-shot setting. In this paradigm, models were prompted with only task instructions, allowing them to function as multi-task systems without task-specific examples. Building upon this foundation, researchers introduced few-shot prompting, wherein carefully designed, high-quality examples included in the prompt enable LLMs to infer reasoning strategies from the provided context~\citep{teaching,jie2023leveraging}. This technique has demonstrated improved performance over zero-shot prompting for certain tasks. However, crafting high-quality demonstrations for a diverse array of reasoning tasks can present practical challenges, and complex few-shot prompts may significantly increase the computational cost during inference.

To further augment the reasoning performance achievable with zero-shot prompting, researchers introduced the influential technique of CoT prompting~\citep{boosting,97,ps,math,cotreasoning,llmzsreasoner,bot,tot,autocot,ghosh2024visual,flux}. Simple instructive phrases, such as ``Let’s think step by step'', guide the model to generate intermediate reasoning steps, which has been shown to improve its problem-solving accuracy. This approach has been subsequently extended by methods like self-consistency~\citep{huang2023large}, which involves generating multiple reasoning paths and selecting the most consistent outcome, thereby enhancing the reliability of the final answer. Nevertheless, the manual design of effective CoT prompts for a wide range of problems can be a labor-intensive process.

Beyond linear reasoning pathways, more advanced prompting strategies have been developed to explore structured reasoning. The tree-of-thoughts (ToT) approach~\citep{tot} models reasoning as a tree-like structure of thought trajectories and employs search algorithms to navigate this structure and explore various potential solution paths. Extending this concept, graph-of-thoughts (GoT)~\citep{GOT} generalizes this structure to a graph, potentially offering more powerful reasoning capabilities and flexible backtracking mechanisms. In recent years, prompting strategies have evolved from straightforward knowledge elicitation to supporting complex, multi-stage reasoning processes, which may or may not incorporate external support.

\section{LLMs’ Mathematical Reasoning from Comprehension and Generation}
\label{sec:comprehension}

Developing large language models (LLMs) capable of sophisticated mathematical reasoning presents a significant challenge. Such models must effectively master abstract symbols, complex notations, and advanced mathematical concepts to solve problems. This endeavor is inherently complex, yet recent years have witnessed considerable advancements in this domain through various techniques and approaches. To navigate this evolving landscape, we propose a structured framework to organize recent innovations. By identifying the core components that underpin mathematical reasoning capabilities, this framework provides a foundation for our subsequent discussion of performance-enhancing techniques.

When addressing mathematical problems, these models first comprehend the relevant knowledge~\citep{allen2024physics} and then proceed to decompose and solve the problem. Inspired by the mechanisms of the human brain~\citep{collins2012reasoning, friederici2011brain}, a structured framework can be conceptualized around two high-level cognitive components: comprehension and answer generation. Comprehension requires LLMs to parse and contextualize mathematical structures and concepts—ranging from arithmetic operations to geometric relationships and theorem formalisms—utilizing both textual and visual representations. Answer generation has evolved from direct prediction to more elaborate step-by-step CoT reasoning.

Responding to mathematical problems via LLMs involves a prompt-guided process that necessitates a thorough understanding of the task and the ability to interpret formal representations. Once the task is comprehended, the model can apply reasoning to derive a solution. For arithmetic problems, this involves performing calculations and manipulating numbers according to established rules. For word problems, it requires extracting mathematical relationships from textual descriptions to formulate and solve mathematical expressions. In geometry, the model must be able to visualize and analyze spatial relationships involving shapes, sizes, angles, and their relative positions. Theorem proving entails leveraging a rich knowledge base, combining formal logical analysis with an understanding of mathematical axioms, theorems, and proof strategies to establish rigorous and sound conclusions. 

\subsection{Diverse Mathematical Representation Formats for LLMs to Comprehend}

\label{sec:Mathematical Representation Formats}
Developing mathematical comprehension in LLMs involves training them on a wide array of mathematical tasks and objectives, designed to address challenges across different domains and levels of difficulty. These tasks span a broad spectrum, from basic arithmetic and algebra typical of primary and middle school curricula, to more advanced topics such as geometry, calculus, and Olympiad-level problems. Furthermore, mathematical problems are presented in multiple formats, including textual descriptions, formal mathematical notation, and visual representations like graphs and geometric figures. By incorporating these diverse problem types into training datasets, LLMs are encouraged to develop a flexible understanding that enables them to process and reason about mathematical concepts across various formats and complexities.

In the early stages of artificial intelligence research~\citep{feigenbaum1963computers, bobrow1964natural}, initial attempts to solve mathematical problems involved designing solver systems reliant on manually written rules and pattern matching. However, these solvers depended heavily on human intervention and could only handle a limited set of predefined scenarios~\citep{slagle1965experiments, fletcher1985understanding}.
Subsequently, semantic parsing-based approaches were introduced~\citep{kwiatkowski2013scaling, goldwasser2014learning}, which aimed to transform problem statements into structured logical representations, akin to syntax trees. These methods, however, sought to explicitly encode human-derived mathematical understanding into models, thereby restricting their applicability to a narrow range of predefined mathematical problems.

\begin{figure}[tbp]
    \centering
    \includegraphics[height=0.35\textheight]{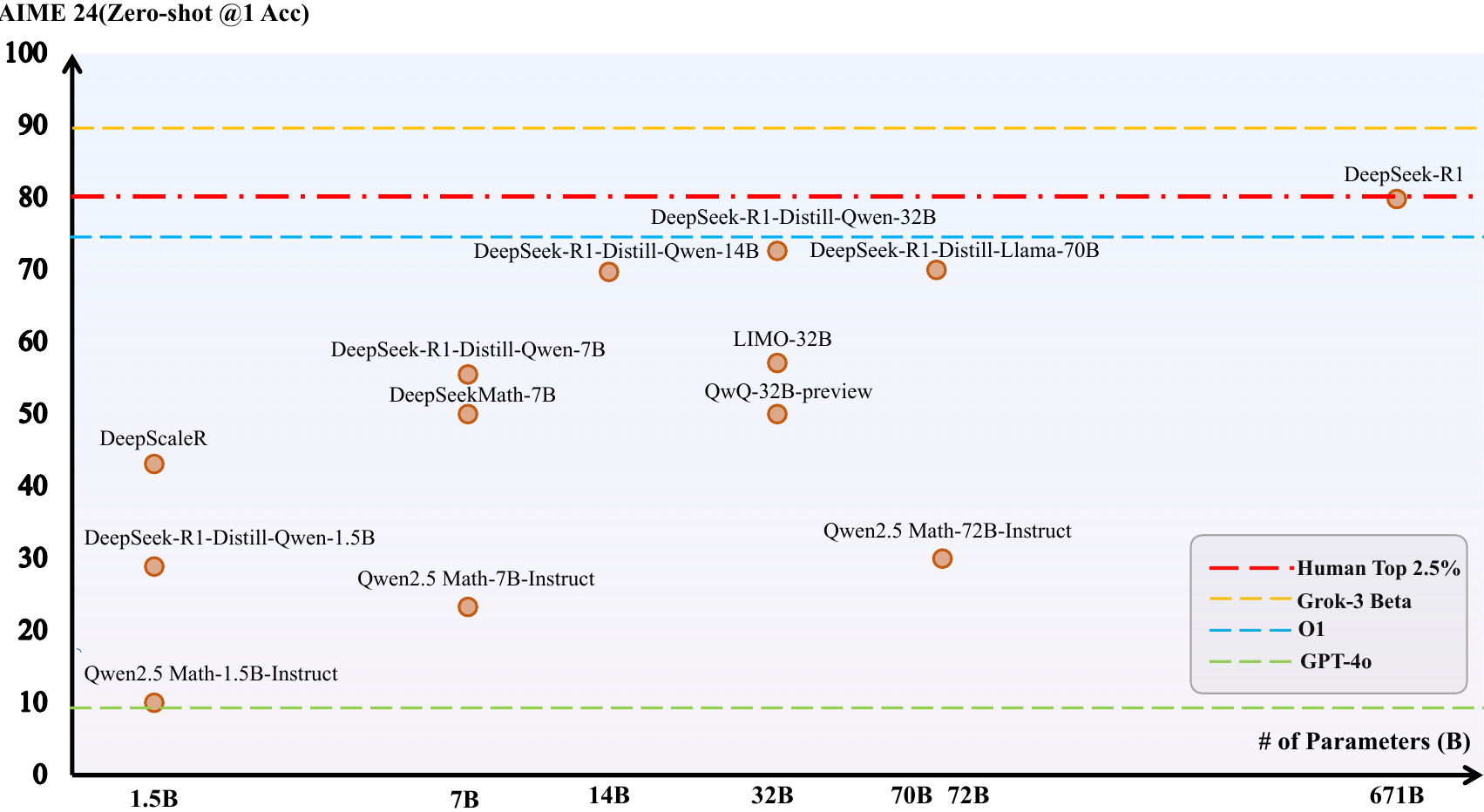}
    \caption{Scaling up LLMs leads to surpassing the top $2.5\%$ of human participants on AIME 2024}
    \label{fig:performance_modelsize}
\end{figure}

With the advent of models like ChatGPT, researchers have observed that formulating models as sophisticated LLMs and scaling up both model and data size can significantly enhance their capabilities~\citep{wei2022emergent}. For example, Figure~\ref{fig:performance_modelsize} illustrates the impact of model scale. Inspired by this observation, training on larger-scale mathematical datasets has been shown to improve a model's performance on mathematical tasks, leading to enhanced comprehension and generalization~\citep{yang2024qwen2, touvron2023llama}. A well-curated dataset is crucial for exposing the model to diverse mathematical contexts and problem-solving patterns, which aids in the development of deeper and more robust mathematical comprehension. For instance, OpenWebMath~\citep{pasteropenwebmath} comprises 14.7B tokens sourced from Common Crawl, offering a broad range of extracted text or \LaTeX ~content that includes core mathematical materials such as theorems, definitions, proofs, questions and answers, and formal mathematics, as well as interdisciplinary documents. MathPile~\citep{wangmathpile} is a high-quality, diverse math-focused corpus containing approximately 9.5B tokens. For a comprehensive list of datasets, please refer to Table~\ref{tab:dataset}.
Adjusting the data distribution within the pretraining corpus, for example by incorporating error-correction data, has been demonstrated to enhance the higher-level reasoning abilities of LLMs~\citep{ye2024physics}.

Furthermore, mathematical reasoning problems frequently involve diverse inputs that extend beyond traditional text-only formats. Over the past year, multimodal mathematical reasoning has emerged as a significant research focus for multimodal large language models (MLLMs).
By introducing additional modalities, such as images, audio, and video, multimodal inputs can reduce the reliance on verbose textual descriptions and provide essential information for reasoning. However, multimodal data inherently exhibit heterogeneity~\citep{liang2024foundations}; specifically, information from different modalities cannot always be directly mapped into a shared latent space. Moreover, compared to text, multimodal inputs often contain a greater amount of noise, such as irrelevant details in images, which imposes additional comprehension challenges~\citep{DBLP:journals/corr/abs-2503-21614}. Therefore, effective multimodal reasoning requires models not only to perceive and understand objects within each modality but also to perform reasoning based on key information embedded in complex multimodal contexts~\citep{DBLP:journals/corr/abs-2503-21614,DBLP:conf/nips/ZhengYTZY23}. As multimodal learning is not the primary focus of this survey, interested readers are referred to \citep{yan2024survey, zhou2025reinforced} for a more comprehensive overview.

Training LLMs on large-scale corpora has become the mainstream approach. Many studies have investigated the internal comprehension mechanisms of LLMs and found evidence of forward planning, where models engage their knowledge and consider multiple possibilities before reacting, and memory shortcuts, where they bypass standard reasoning paths~\citep{lindsey2025biology}. By defining model features as human-interpretable concepts, ranging from low-level elements (e.g., specific words or phrases) to high-level abstractions (e.g., sentiment, plans, reasoning steps), research has shown that key concepts in a prompt are actively represented and activated inside LLMs~\citep{templeton2024scaling}. Furthermore, LLMs have demonstrated the ability to identify the underlying structure behind surface-level problems and invoke distilled skills to solve associated tasks~\citep{guo2024learningpatternmatchingassaying, didolkar2024metacognitivecapabilitiesllmsexploration}. Also, semantic embeddings in large language models have been shown to exhibit linear structure, enabling concept relationships to be captured through vector arithmetic~\citep{arora2019latentvariablemodelapproach}. Collectively, these findings suggest that LLMs go beyond surface-level pattern matching; they internalize conceptual structures and flexibly integrate them across diverse reasoning scenarios.

\subsection{Chain-of-Thought: A Key Driver of LLM Generation}
\label{subsubsec:prompting-cot}
LLMs encode task instructions into latent representations through transformer architectures, subsequently generating responses via autoregressive sequence completion. However, direct generation methods frequently fail to produce accurate solutions for tasks requiring complex mathematical reasoning. CoT prompting addresses this limitation by guiding LLMs to generate intermediate reasoning steps before producing final answers, thereby enhancing performance on multi-step problems. This section presents a characterization of CoT and examines the underlying mechanisms that account for its effectiveness.

\subsubsection{Formulation of CoT}
CoT prompting induces LLMs to produce intermediate reasoning steps prior to generating final answers. This approach facilitates progressive complexity reduction, information augmentation, and irrelevant information filtering throughout the reasoning process. The demonstrated effectiveness of CoT has catalyzed the development of advanced techniques for enhancing LLM reasoning capabilities. 
\citet{cotreasoning} demonstrated that generating sequences of intermediate reasoning steps significantly improves LLM performance on complex reasoning tasks, particularly those involving mathematical or logical operations~\citep{sprague2024cot}. Furthermore, \citet{llmzsreasoner} demonstrated that LLMs can perform zero-shot CoT reasoning through the simple addition of reasoning-trigger phrases (e.g., ``Let's think step by step'') to prompts, enabling reasoning capabilities without explicit step-by-step demonstrations.

\begin{figure}
    \centering
    \includegraphics[width=0.9\linewidth]{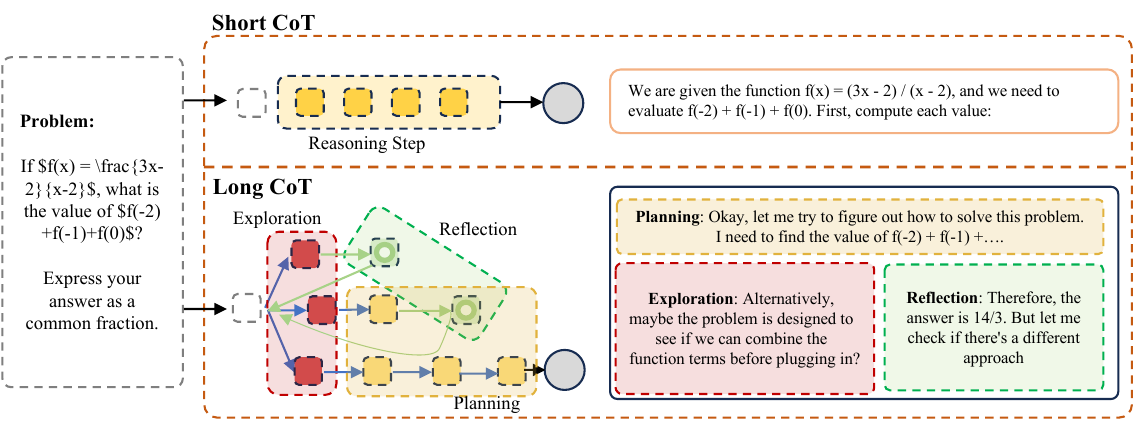}
    \caption{An illustration of short and long CoT.}
    \label{fig:short_long}
\end{figure}

Initial CoT implementations employ shallow, linear reasoning processes characterized by sequential answer derivation and limited intermediate steps~\citep{gsm2025seyed}. Recent advances, including OpenAI O1~\citep{openai2024o1} and DeepSeek R1~\citep{guo2025deepseek}, introduce extended sequential CoT reasoning through test-time scaling, enabling more comprehensive and structured reasoning processes. This extended approach, termed \emph{long CoT} \citep{longcot, li2025surveyllmtesttimecompute}, incorporates iterative exploration and self-reflection within problem spaces. Figure~\ref{fig:short_long} illustrates the distinction between long CoT and traditional short CoT approaches. Test-time scaling enables models to identify inconsistencies in intermediate steps and implement corrective measures to maintain coherence and accuracy. Additionally, models may explore multiple solution paths and backtrack when specific approaches prove incorrect, yielding more robust and reliable inference outcomes.

\subsubsection{Unveiling the Mechanism behind CoT}
While CoT demonstrably enhances LLM performance in mathematical domains~\citep{cotreasoning, llmzsreasoner}, the underlying mechanisms remain incompletely understood. A comprehensive investigation of these fundamental factors is essential for maximizing CoT effectiveness. This review examines mechanisms from both theoretical and empirical perspectives.

\begin{table}[htbp]
\caption{A summary of pre-train math datasets. \textbf{Synth.} indicates that the dataset contains synthetic data.}

\setlength{\extrarowheight}{2pt}
\footnotesize
\centering
\begin{tabular}{l|llll}
\toprule

\rowcolor{gray!20}\textbf{Dataset} & \textbf{Target Domain} & \textbf{Synth.} &\textbf{Size(Tokens/Pairs)}& \textbf{Release Time} \\ 
\midrule
AMPS~\citep{hendrycks2measuring} & Math Competition & \ding{51} & 0.7B  & Mar-2021\\ 

\rowcolor{gray!10}ProofPile~\citep{proofpile} & General Math & \ding{51} & 8.3B  & Nov-2022\\ 
ProofPile2~\citep{azerbayevllemma} & General Math & \ding{51} & 55B  & Oct-2023\\ 
\rowcolor{gray!10}OpenWebMath~\citep{pasteropenwebmath} & General Math & \XSolidBrush &  14.7B & Oct-2023\\ 
MathPile~\citep{wangmathpile} & General Math & \ding{51} &  9.5B & Dec-2023\\ 

\rowcolor{gray!10}AutoMathText~\citep{zhang2024autonomous} & General Math & \XSolidBrush &  - & Feb-2024 \\ 

\bottomrule
\end{tabular}
\label{tab:dataset}
\end{table}

From theoretical perspective,
~\citet{feng2023towards} employed circuit complexity theory~\citep{sanjeev2009compu} to demonstrate that while bounded-depth Transformers require super-polynomial size for direct solution of basic arithmetic and equation tasks, autoregressive Transformers achieve successful solutions through CoT derivation generation with constant-size architectures. These models generalize effectively to longer input sequences, indicating that CoT enables internalization of reasoning processes rather than mere memorization of input-output mappings. From an expressiveness perspective, CoT enhances transformer capacity by enabling intermediate token generation before final answer production. This methodology provides greater expressive power for sequential reasoning tasks, as intermediate tokens function as recurrent states~\citep{merrill2024the}. 
\cite{li2024chain} formulated and analyze the hypothesis that CoT enables serial computations beyond the capabilities of vanilla transformers. \citet{ben2023why} applied a Bayesian framework to elucidate how intermediate steps contribute to enhanced reasoning performance, while \citet{rasual2023why} developed a two-level hierarchical graphical model characterizing LLM reasoning sequence generation. From an in-context learning perspective, CoT improves performance by structuring compositional function learning into two phases, reducing sample complexity and enabling complex function learning beyond non-CoT method capabilities~\citep{li2023dissecting}.

From the empirical perspective, \citet{DBLP:journals/corr/abs-2307-13339} demonstrated that CoT enables LLMs to maintain robust attention on semantically relevant prompt tokens. \citet{DBLP:conf/acl/WangM0S0Z023} identified two critical factors affecting CoT effectiveness: semantic relevance of demonstration examples and correct ordering of reasoning steps. \citet{DBLP:conf/acl/JinYSZHMZD24} established a task-dependent relationship for optimal CoT length, where simpler tasks benefit from shorter sequences while complex tasks require extended reasoning steps.
Despite evidence supporting CoT superiority over direct answer generation, empirical observations reveal that language models frequently produce correct answers despite errors in intermediate reasoning. Competition-level tasks exhibit error rates within generated processes reaching 51.8\%~\citep{zheng2024processbench}. \citet{sprague2024cotcotchainofthoughthelps} demonstrated that CoT responses may be suboptimal for non-symbolic reasoning tasks. Recent investigations indicate that LLM-generated reasoning steps often lack reliability and fail to accurately reflect step-by-step logical processes~\citep{arcuschin2025chain, chenreasoning}. Current LLM-generated CoT does not constitute rigorous chains of logically entailed steps but rather resembles heuristic processes that leverage recursive CoT structure to generate sufficiently rich intermediate content, increasing effective model depth and supporting final answers despite intermediate step errors.

\section{Methods for Boosting Reasoning}
\label{sec:boosting}

During the pre-training phase, LLMs are trained on vast amounts of data to enhance the comprehension of mathematical problems. However, pre-trained models often struggle with producing contextually appropriate responses. Therefore, improving model performance is crucial for effective mathematical reasoning. Fine-tuning plays a crucial role in enhancing a model’s instruction-following and generation capabilities. 

\subsection{Supervised Fine-Tuning}
\label{subsec:SFT}
Supervised Fine-Tuning (SFT) aims to align a pretrained language model with high-quality, human-crafted by supervised learning. These datasets follow clear formatting patterns, which provide structural priors that help constrain and guide the model’s generation space. This structure plays a crucial role in enabling effective exploration during subsequent RL. The efficacy of SFT hinges on the construction of high-quality supervised demonstrations. 

\subsubsection{Constructing High-Quality Demonstrations}
\label{subsubsec:highquality SFT}

In SFT, the quality and structural characteristics of training data fundamentally determined the behavioral patterns and computational capabilities of language models. As SFT directly modified model parameters through demonstration-based learning, model performance exhibited high sensitivity to the diversity, relevance, and correctness properties of fine-tuning datasets. The construction of high-quality instructional data that accurately reflected target applications and alignment objectives constituted a critical requirement for successful model adaptation. This section examines algorithmic approaches to data construction and curation for reasoning tasks. A comprehensive overview of relevant datasets is provided in Table~\ref{tab:dataset_sft}.

\begin{table}[tbp]

\caption{A summary of SFT math datasets. \textbf{Synth.} indicates that the dataset contains synthetic data.}

\footnotesize
\centering
\begin{tabular}{l|llll}
\toprule

\rowcolor{gray!20}\textbf{Dataset} & \textbf{Target Domain} & \textbf{Synth.} &\textbf{Size(Tokens/Pairs)}& \textbf{Release Time} \\ 
\midrule


NaturalProofs~\citep{welleck1naturalproofs} & Theorem Proving & \XSolidBrush &  48K & Mar-2021\\
Lean Workbook~\citep{ying2024lean} & Theorem Proving & \XSolidBrush &  57K & Jun-2024\\  
NuminaMath~\citep{li2024numinamath} & Math Competition & \ding{51} &  860K & Jul-2024\\ 

CARP~\citep{zhang2024evaluating} & General Math (zh) & \XSolidBrush &  10M & Jun-2023\\ 
 MetaMathQA~\citep{yumetamath} & General Math & \ding{51} &   395K & Sep-2023\\ 
MathInstruct~\citep{mammoth} & General Math & \ding{51} &  260K & Sep-2023\\
 MMIQC~\citep{augmenting} & General Math & \ding{51} &  1.57M & Jan-2024\\ 
OpenMathInstruct-
1~\citep{toshniwal2024openmathinstruct} & General Math & \ding{51} &  1.8M & Feb-2024\\ 
MathScaleQA~\citep{tang2024mathscale} & General Math & \ding{51} &  2M & Mar-2024\\ 
\rowcolor{gray!10}WebInstruct~\citep{yue2024mammoth2} & General Math & \XSolidBrush
&  10M & May-2024\\ 
OpenMathInstruct-2~\citep{toshniwal24openmathinstruct} & General Math & \ding{51} &  14M & Oct-2024\\ 

\bottomrule
\end{tabular}
\label{tab:dataset_sft}
\end{table}

Recent research by \citep{ho2023large, magister2023teaching, guo2025deepseek} demonstrated that constructing training data using strong LLMs significantly improved the mathematical reasoning capabilities of smaller LLMs. OpenMathInstruct-1~\citep{toshniwal2024openmathinstruct} augmented synthetic data generation through code-interpreter solutions produced by GPT-4. \citet{mammoth} utilized GPT-4 to generate Program-of-Thought (PoT) rationales, thereby enhancing tool usage capabilities. Beyond simple answer expansion, leveraging LLM knowledge for question bootstrapping enhanced problem coverage. \citep{liu2023tinygsm, li2024common} employed strong LLMs to generate semantically similar questions and corresponding answers. However, the generated questions exhibited limited diversity due to textual and conceptual similarity constraints. \citet{li2024mugglemath} and \citet{yang2024qwen2} expanded question sets through diverse modification techniques, including numerical alterations, conceptual modifications, and complexity augmentation.

\citet{yumetamath} bootstrapped mathematical question generation through multiple techniques: Rephrasing, Self-Verification~\citep{weng2023large}, FOBAR~\citep{jiang2024forward}, and answer augmentation. \citet{huang2024key} extracted key concepts from existing datasets and utilized these concepts alongside original problems as guidelines for generating novel questions. \citet{bansal2024smaller} demonstrated that data generated by smaller models exhibited greater distributional diversity. \citet{ding2024unleashing} employed smaller models to generate questions de novo without seed data dependencies, utilizing complex augmentation constraints. \citet{adarsh2024siked} ensured diversity by combining generative outputs from multiple smaller models, while \citet{li2025preserving} actively preserved diversity in fine-tuned models through targeted methodologies.

Beyond high-quality solution generation, datasets incorporating erroneous reasoning enabled LLMs to develop error detection and correction capabilities essential for advanced mathematical reasoning. \citet{an2023learning} enhanced mathematical reasoning through the incorporation of error-correction data during the fine-tuning phase. This error-correction data, generated by GPT-4, included error identification, correction processes, and final answer generation. \citet{liang2023let} employed a teacher LLM to identify weaknesses in student LLMs and generated targeted problems for training dataset augmentation. \citet{he2023teacherlm} provided comprehensive training signals including reasoning processes, foundational knowledge, and common error patterns during answer generation. \citet{dai2024beyond} generated dual chains of thought encompassing both correct and incorrect reasoning paths from teacher models, utilizing minimum edit distance to identify critical reasoning steps. This approach emphasized learning fundamental reasoning mechanisms rather than superficial fine-tuning.

\subsubsection{Constructing Demonstrations in Long Chain-of-Thought Format}
\label{subsec:template}
Recent studies demonstrated that enabling LLMs to generate extended CoT sequences during test-time inference significantly enhanced reasoning accuracy~\citep{brown2024large, snell2024scaling}. Fine-tuning methodologies increasingly adopted the long CoT paradigm, wherein allocating additional computational resources to CoT reasoning during both training and inference phases yielded consistent performance improvements. Through the construction and utilization of long CoT demonstrations, supervised fine-tuning enabled LLMs to acquire the capability of generating extended CoT outputs that exhibited diverse reasoning processes, including interactive exploration and self-reflection mechanisms.

Deepseek R1~\citep{guo2025deepseek}, extending Deepseek R1 Zero, collected high-quality cold-start data and implemented a structured Markdown format. The system defined output format as $\langle response_process \rangle~\langle summary \rangle$, wherein reasoning processes preceded summary generation of reasoning paths. This architectural design enhanced output readability and interpretability. Kimi K1.5~\citep{team2025kimi} utilized a high-quality long CoT dataset, employing SFT as a warmup phase that improved the generation of logically coherent and detailed responses. LIMO~\citep{ye2025limo} and s1~\citep{muennighoff2025s1} challenged the necessity of large sample sizes, demonstrating that minimal sample sets successfully activated reasoning capabilities in foundational LLMs. Satori~\citep{DBLP:journals/corr/abs-2502-02508} introduced a critic model for constructing multi-step demonstrations with reflection mechanisms, facilitating enhanced multi-step reasoning capabilities in trained models.


\subsection{Reinforcement Learning}
\label{subsec:RL}
Reinforcement learning has been employed to enhance model generation capabilities through CoT reasoning. Prior to generating final answers $y$, models produced intermediate CoT reasoning steps $z\sim \pi_\theta(\cdot| x)$. Early reinforcement learning implementations in mathematical reasoning focused on optimizing standard CoT generation through outcome reward models and process reward models. Recent developments have adapted reinforcement learning techniques to optimize long CoT generation, primarily utilizing rule-based rewards to ensure coherent extended reasoning trajectories. The fundamental RL approach remains consistent across both applications, with the primary distinction lying in the reward design and the target output format-whether optimizing for standard CoT or extended long CoT reasoning processes.

\subsubsection{Reward Modeling for Reasoning}
\label{subsec:reward}

In the application of reinforcement learning to improve LLMs, the reward model constitutes a critical component. A reward model assigns a numerical score $r_{\theta}(x,y) \in [0,1]$ to estimate the probability that a solution $y$ or intermediate step is correct for a given problem $x$. Reward models can be categorized into three primary types: outcome reward models (ORM), process reward models (PRM), and rule-based reward systems. This section examines each category and their respective contributions to reinforcement learning training stability.

\paragraph{Outcome Supervised Reward}
The standard approach for training ORMs in reasoning tasks involved fine-tuning an LLM as a classifier on datasets containing correct and incorrect solutions. These datasets were annotated either by human evaluators or generated from frozen LLMs during self-improvement processes, utilizing binary cross-entropy loss~\citep{cobbe2021training, stepawareverifier}. Given a reward-modeling dataset $\mathcal{D}_{\mathrm{RM}} = \mathcal{D}_{\mathrm{incorrect}} \cup \mathcal{D}_{\mathrm{correct}}$, discriminative reward models were trained according to:

\[
\mathcal{L}(\theta, \mathcal{D}_{\mathrm{RM}})
= - \mathbb{E}_{(x,y^+) \sim \mathcal{D}_{\mathrm{correct}}} 
\bigl[\log r_\theta(x,y^+)\bigr]
\;-\;
\mathbb{E}_{(x,y^-) \sim \mathcal{D}_{\mathrm{incorrect}}}
\Bigl[\log\bigl(1 - r_\theta(x,y^-)\bigr)\Bigr],
\]

where $r_\theta(x,y) = \sigma(z_{\mathrm{cls}})$ and $z_{\mathrm{cls}} = \mathrm{logit}_\theta(\mathrm{cls} \mid y,x)$. Here, $y^+$ denoted correct solutions, $y^-$ denoted incorrect solutions, and \texttt{cls} corresponded to a special vocabulary token. \citet{zhang2024generative} employed a balanced data mixture between correct $(\mathcal{D}_{\mathrm{correct}})$ and incorrect $(\mathcal{D}_{\mathrm{incorrect}})$ problem-solution pairs. \citet{treebased} proposed a methodological shift from binary classification loss to preference-based loss for verifier training.

Multiple studies addressed the reduction of human annotation requirements through the LLM-as-a-verifier approach, wherein off-the-shelf LLMs evaluated solutions via prompting~\citep{madaan2024self, generateseqselfcorret, zhang2024llama}. These LLMs assigned rewards or penalties to both outcomes and intermediate steps based on reasoning quality and alignment with predefined criteria. While these methods demonstrated effectiveness in language tasks, their performance in mathematical problem-solving remained limited. \citet{zhang2024generative} proposed GenRM, which outperformed discriminative verifiers by integrating CoT reasoning into the verification process.

In applications, ORMs evaluated responses at generation completion to guide policy optimization. DeepseekMath~\citep{shao2024deepseekmath} leveraged an ORM to optimize policies using the GRPO algorithm, while Qwen2.5-Math~\citep{yang2024qwen2} employed a hybrid reward mechanism combining ORM with rule-based rewards to enhance training stability and performance.

\paragraph{Process Supervised Reward}
Recent investigations by \citet{lightman2023let} indicated that PRMs outperformed ORMs in reasoning tasks. PRMs $(P \times S \to \mathbb{R}^+)$ assigned scores to individual reasoning steps within solution $s$, typically trained using:

\[
\mathcal{L}_{\text{PRM}} \;=\; \sum_{i=1}^{K} \Bigl( 
y_{s_i}\,\log r_{s_i} \;+\; \bigl(1 - y_{s_i}\bigr)\,\log\bigl(1 - r_{s_i}\bigr)
\Bigr)
\]

where $y_{s_i}$ represented the ground truth label for step $s_i$ (the $i$-th step of $s$), $r_{s_i}$ denoted the sigmoid score assigned by the PRM to step $s_i$, and $K$ indicated the total number of reasoning steps in $s$.

Given the labor-intensive nature of process reward annotation, \citet{zhangmcts*} developed methods for learning process rewards through final reward guidance. REFINER~\citep{refiner} provided structured feedback on reasoning errors through intermediate step evaluation. \citet{provers} trained process reward models to measure the likelihood of future correct response generation by incorporating additional policies. TSMC utilized intermediate target distributions for resampling during Monte Carlo processes~\citep{twistmcts}. \citet{luo2024improve} proposed a divide-and-conquer style MCTS algorithm for efficient collection of high-quality process data.

Compared to outcome rewards, process rewards provided more detailed feedback, demonstrating greater potential to enhance generators~\citep{PRMgivebetterrm}. \citet{wang2024math} utilized automatically constructed PRMs to supervise LLMs through step-by-step PPO. Implicit PRM~\citep{yuan2024implicitprm} extended ORM training by implicitly learning process labels without requiring additional annotations. \citet{stepkto} introduced a training framework combining process-level and outcome-level binary feedback to guide LLMs toward more reliable reasoning trajectories.

\paragraph{Rule-Based Reward}
In mathematical reasoning contexts, rule-based rewards were designed based on the verifiability of final answers and intermediate reasoning steps. The reward function evaluated whether the final answer exactly matched the ground truth solution, returning a binary reward: $r=R(y^*,y)$, where $r = 1$ if and only if the model's final answer exactly matched the ground truth $y^*$. Format rewards could additionally be incorporated to distinguish reasoning paths from final answers~\citep{guo2025deepseek}.

Recent studies~\citep{guo2025deepseek, DBLP:journals/corr/abs-2502-02508, deepscaler2025, yue2025does} increasingly focused on rule-based rewards due to their provision of accurate and reliable reward signals that supported stable reinforcement learning training. Compared to learned reward models, rule-based rewards mitigated issues such as reward hacking by providing deterministic feedback grounded in task-specific logic. ReFT~\citep{luong2024reft} explored rule-based RL with SFT warm-up, achieving significantly superior performance compared to SFT alone in mathematical domains. Deepseek R1~\citep{guo2025deepseek} employed a rule-based reward function and achieved continued improvement beyond 8,000 training steps, ultimately attaining performance comparable to or exceeding OpenAI's o1.

\subsubsection{Reinforcement Learning}
\label{subsec:rl}
Inspired by reinforcement learning from human feedback (RLHF), reinforcement learning was introduced for mathematical reasoning enhancement in LLMs~\citep{luong2024reft, kazemnejad2024vineppo, gehring2024rlef, provers, li2024rl}. While supervised fine-tuning relied on offline datasets and exhibited susceptibility to compounding errors due to distributional shifts, RL mitigated these limitations through online, reward-driven optimization. In response evaluation frameworks, outcome rewards assessed entire responses by evaluating final outcome confidence levels. Process rewards provided scores at each reasoning step conclusion, delivering more comprehensive and informative supervision signals.

\paragraph{LLM-Specific RL Algorithms}
\label{subsubsec:other-rl}

Proximal Policy Optimization (PPO)~\citep{ppo} emerged as the predominant RL method for RLHF following its adoption in ChatGPT. However, as a general-purpose RL algorithm, PPO required an additional critic model, substantially increasing computational costs and GPU memory consumption. To address these limitations, ReMax~\citep{liremax} pioneered the elimination of the critic model by recognizing that reinforcement learning for LLMs exhibited simpler properties than general reinforcement tasks. ReMax leveraged the REINFORCE algorithm and introduced greedy sampling responses to calculate reward baselines, maintaining training stability without the computational overhead of a critic model.

Building upon this approach, RLOO~\citep{rloo} similarly eliminated the critic model but employed Monte Carlo sampling for improved baseline estimation. GRPO~\citep{shao2024deepseekmath} adopted the PPO objective while calculating advantages from group-normalized rewards, achieving comparable performance without the critic model overhead. Reinforce++~\citep{hu2025reinforce++} employed similar techniques with additional optimizations for training efficiency. Furthermore, addressing the challenge of rapidly decreasing policy entropy during training, which limited exploration capabilities, DAPO~\citep{yu2025dapo} introduced the Clip-Higher strategy, building upon GRPO to maintain exploration throughout the training process.

\paragraph{Reinforcement Learning for Long CoT}

Long CoT reinforcement learning represented a paradigm shift in optimizing LLMs by leveraging RL with rule-based rewards to enhance extended reasoning capabilities. While OpenAI's o1 model demonstrated the effectiveness of allocating increased computational resources during both training and inference, implementation details remained proprietary. DeepSeek R1 achieved comparable or superior performance through a purely RL-based approach, catalyzing renewed research interest in long CoT RL methodologies. The fundamental breakthrough involved emulating System 2-style reasoning through three critical factors:

\begin{itemize}
    \item \textbf{Golden Reward}: Previous vanilla RL methods frequently encountered reward model saturation, wherein reward model inaccuracies induced reward hacking behaviors. DeepSeek R1 addressed this limitation by employing rule-based rewards determined through comparison with ground truth answers, enabling more reliable correctness evaluation and stable exploration dynamics.
    
    \item \textbf{Scaling CoT Length}: Extending CoT from short to long sequences enhanced the model's exploration capabilities and strengthened intrinsic reasoning abilities. Increased generation length enabled exploration of diverse response spaces and facilitated the emergence of advanced capabilities including verification and reflection mechanisms, promoting deeper analytical thinking.
    
    \item \textbf{Pure RL Training}: DeepSeek R1-zero implemented pure RL training by bypassing the SFT phase entirely, directly applying reinforcement learning to the base model. This approach enabled exploration of broader distributional spaces beyond the constraints imposed by SFT datasets, thereby maximizing exploration potential~\citep{zeng2025simplerl}.
\end{itemize}

Regarding reward mechanism design, R1 experimented with PRM integration but encountered intermediate step labeling challenges. Alternative approaches emerged: Kimi K1.5~\citep{team2025kimi} utilized ORM and GenRM~\citep{zachary2024critique, bo2024critics} for reward generation, continuing training after SFT with long CoT data to achieve performance comparable to o1. Satori~\citep{DBLP:journals/corr/abs-2502-02508} explicitly integrated verifier and search capabilities within a unified model architecture, internalizing reasoning mechanisms to enhance problem-solving efficiency. Self-Reward methodologies~\citep{weiselfrewarding, pang2023yang} incorporated self-critique and verification capabilities through SFT, subsequently enhancing these abilities via RL to enable robust reasoning and self-improvement during inference. RFTT~\citep{Zhang2025RFTT} integrated multiple cognitive capabilities through specialized tokens such as $\langle analyze \rangle$, $\langle verify\rangle$, and $\langle refine \rangle$, enabling structured long CoT construction.

\subsubsection{Direct Preference Optimization}
\label{subsec:dpo}

To address the complexity of online RL optimization, Direct Preference Optimization (DPO) \citep{DPO} was proposed as an alternative approach that directly utilized offline pair-wise preference data for model optimization. This methodology significantly streamlined the training pipeline and enhanced training stability. Unlike traditional methods such as RLHF, which required training a reward model followed by policy model optimization, DPO eliminated the necessity for separate reward model training, substantially simplifying the training process. DPO demonstrated particular effectiveness in optimizing language models through alignment with human preferences, facilitating more efficient model fine-tuning \citep{llama3}.

\citet{richard2024iteractive} modified the DPO loss function by introducing an additional negative log-likelihood term. Through iterative optimization of reasoning paths via comparative evaluation of winning and losing pairs, this approach achieved improvements in mathematical reasoning tasks. SimPO~\citep{meng2024simpo} incorporated average log probability calculations, effectively aligning with model generation while eliminating dependency on reference models, thereby improving computational and memory efficiency.

In the context of process-level optimization, Step-DPO~\citep{lai2024step} extended DPO by specifically targeting long-chain reasoning, producing notable improvements in mathematical word problem solving. Additionally, \citet{tokendpo} modified DPO to optimize policies at the token level, further refining LLM alignment with human preferences.

\paragraph{Limitations of DPO.} Despite its widespread adoption as an alternative to RLHF, DPO fundamentally operates within a supervised learning paradigm\footnote{Here we categorize DPO as a supervised learning paradigm rather than reinforcement learning or offline reinforcement learning, as it fundamentally operates through likelihood maximization (and minimization for negative samples) over fixed preference datasets.} rather than a true reinforcement learning framework, resulting in several critical limitations. First, DPO lacks the exploration capabilities inherent to RL methods, as it cannot utilize out-of-preference data or policy-generated samples to discover novel high-quality responses beyond the training distribution~\citep{li2024rl, xu2024dpo}. Second, theoretical analyses revealed that DPO exhibits asymmetric optimization dynamics, decreasing the probability of dispreferred responses faster than increasing preferred ones, which explains its sensitivity to SFT quality and tendency to hinder learning capacity~\citep{feng2024towards}. Third, empirical studies demonstrated that DPO's implicit reward model generalizes poorly under distribution shifts~\citep{li2024rl, lin2024limited}, with accuracy drops reaching 7\% in out-of-distribution settings compared to explicit reward models~\citep{lin2024limited}. These collective findings indicate that while DPO offers computational efficiency advantages, it sacrifices the exploratory benefits, generalization capabilities, and adaptive flexibility that characterize true reinforcement learning approaches.


\subsection{Test Time Inference with Structural Search}
\label{subsec:testing-search}

\begin{figure}[htbp]
    \centering
    \includegraphics[width=0.9\linewidth]{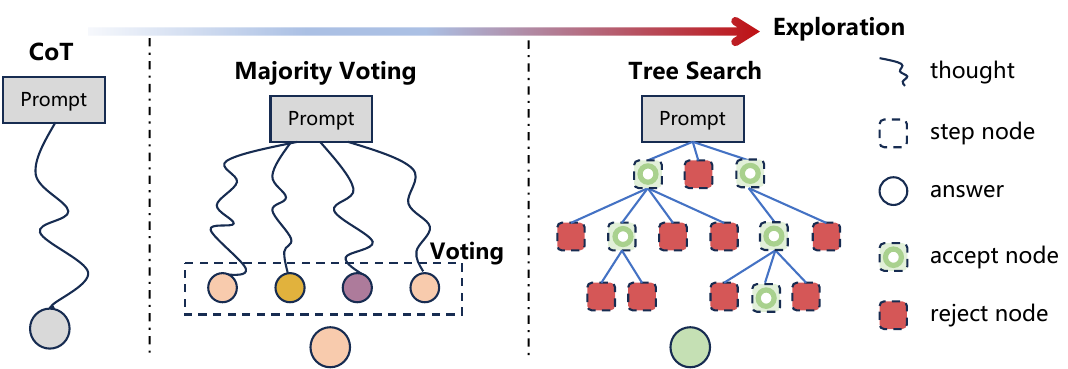}
    \caption{Schematic illustrating of different method.}
    \label{fig:cot2tree}
    \vspace{-2mm}
\end{figure}

Test-time scaling through increased generation burden has been shown to enhance model mathematical reasoning capabilities~\citep{wu2024inference, snell2024scaling}. While majority voting~\citep{wangself} enables answer derivation through extensive data collection, complex mathematical problems with high token budgets necessitate tree search methods for exhaustive solution space exploration. The distinction between these approaches is illustrated in Figure~\ref{fig:cot2tree}.

In the decision-making process of LLMs, structured tree search integration enhances planning through systematic exploration. This approach enables LLMs to concurrently maintain and investigate multiple potential solutions during generation, dynamically assessing current states while strategically planning ahead or backtracking to refine reasoning paths. \citet{tot} proposed the Tree-of-Thoughts (ToT) framework, integrating depth-first search (DFS) and breadth-first search (BFS) to enable systematic exploration and backtracking over intermediate reasoning steps. \citet{besta2024graph} extended ToT by introducing the Graph of Thoughts (GoT) structure, supporting cyclic and interconnected reasoning steps that capture complex dependencies and enable iterative refinement processes. Forest-of-Thought (FoT)~\citep{fot2024zhen} further advanced this paradigm by integrating multiple reasoning trees, facilitating collective decision-making for solving complex mathematical problems with enhanced reliability and efficiency.

However, these methods typically relied on heuristic or simplistic search strategies, potentially limiting efficiency in large search spaces. To address this limitation, Monte Carlo Tree Search (MCTS) emerged as a powerful decision-making algorithm for enhanced search efficiency. MCTS constructs structural trees for optimal path exploration, where each node represents a reasoning state~\citep{browne2012survey, chaslot2008monte}. During tree expansion, future states undergo simulation to identify valuable nodes, followed by value function updates through the backup process to refine node evaluations~\citep{coulom2006efficient}.

Recent research demonstrated successful MCTS integration with LLMs. Reasoning-via-Planning (RAP)~\citep{hao2023reasoning} integrated MCTS to enhance reasoning by repurposing LLMs as both world models for state prediction and reasoning agents for action generation. AlphaMath~\citep{chen2024alphamath} employed MCTS during inference-time reasoning, integrating value models with LLMs to autonomously generate process supervision and step-level evaluation signals within MCTS rollouts. LLaMA-Berry~\citep{zhang2024llama} combined MCTS with iterative Self-Refine~\citep{madaan2024self}, dynamically optimizing reasoning paths through exploration and self-critique, guided by pairwise preference reward models (PPRM) that globally evaluated solution quality via Enhanced Borda Count—a method combining basic Borda Count algorithms with transitive closure of preferences computed using the Floyd-Warshall algorithm~\citep{warshall1962theorem}.

Beyond inference-time applications, MCTS integration during training improved exploration efficiency. \citet{feng2023alphazero} proposed TS-LLM, leveraging learned value functions and AlphaZero-like algorithms to guide LLMs during training. rStar-Math~\citep{guan2025rstar} utilized MCTS for generating candidate reasoning steps, guided by process preference models evaluating step-wise quality through multi-turn training.

\subsection{Self-Improvement}
\label{subsubsec:selfimprove}
The construction of extensive rationale datasets for inducing language model rationale generation in downstream tasks presents significant resource constraints in terms of both time and cost. Self-improvement methodologies have emerged as a viable alternative, leveraging the inherent capabilities of LLMs to refine CoT reasoning processes. A fundamental challenge in this approach concerns the generation and selection of high-quality responses to ensure methodological reliability and effectiveness.

During the inference phase, several approaches have been developed to enhance response quality. Self-Refine~\citep{madaan2024self} implements an iterative refinement mechanism utilizing feedback loops. Self-Verification~\citep{weng2023large} incorporates answer validation procedures to mitigate errors in generated responses. RISE~\citep{qu2024Recursive} facilitates inference-time self-improvement through fine-tuning on distilled recursive introspection data.

Contemporary research has systematically investigated the enhancement of LLM capabilities through fine-tuning on self-synthesized responses, which can be categorized according to two distinct methodological approaches. The first approach emphasizes the generation of superior solution quality. STaR~\citep{zelikman2022star} employs few-shot prompting with ground truth answers to generate enhanced rationales. Quiet-STaR~\citep{zelikman2024quiet} implements token-level thought rationale generation to improve prediction accuracy. ReST$^{EM}$~\citep{avi2024beyond} applies the Expectation-Maximization algorithm~\citep{anthony2017thinking} for iterative answer quality refinement. Both rStar-Math~\citep{guan2025rstar} and ReST-MCTS$^*$~\citep{zhangmcts*} incorporate Monte Carlo Tree Search to augment exploration capabilities. SIRLC~\citep{pang2023yang} utilizes the LLM as an intrinsic reward model for reinforcement learning-based training. ReGenesis~\citep{peng2024regenesis} constructs abstract-to-concrete response pathways through self-synthesized guidelines, demonstrating robust generalization to out-of-distribution tasks. Auto-CEI~\citep{zhao2024automatic} employs expert iteration~\citep{anthony2017thinking} for near-policy reasoning path exploration, implementing error correction mechanisms to minimize cumulative errors.

The second approach addresses the selection of high-quality responses from generated candidates. \citet{huang2023large} applies self-consistency methods~\citep{wangself} for high-confidence response identification. V-STaR~\citep{hosseini2024v} implements a DPO-based verifier for solution correctness assessment. Both rStar-Math and ReST-MCTS$^*$ incorporate process reward guidance to ensure step-level error mitigation and reasoning trajectory reliability.
\subsection{Utilizing External Knowledge}
\label{subsec:RAG}
Large language models, despite demonstrating substantial reasoning capabilities, exhibit systematic limitations in tasks requiring precise computational accuracy or access to real-time information, frequently manifesting as hallucinations or factual inaccuracies. Recent methodological developments have addressed these constraints through two principal approaches: external tool integration and retrieval-augmented generation (RAG).

External tools comprise specialized functional modules accessed through standardized APIs or web interfaces, including web search engines, symbolic computation systems, and code execution environments. These tools extend model capabilities beyond static parameter encoding by enabling real-time information retrieval, precise mathematical computation, and dynamic program execution. In contexts involving complex mathematical expressions or temporally-sensitive queries, LLMs demonstrate increased susceptibility to generating plausible yet factually incorrect responses, a limitation that external tool integration effectively addresses. \citet{mathtool} developed a template-based decision framework for tool invocation timing, though this approach lacks support for sequential or interdependent tool utilization. \citet{toolalpacageneralizedtoollearning} implemented a self-instruct paradigm for diverse tool-use scenario generation, similarly constrained by limited multi-tool coordination capabilities. \citet{mathtool} established a comprehensive framework integrating knowledge retrieval (Bing Web Search), program generation and execution (Python), and symbolic computation (WolframAlpha API), demonstrating measurable performance improvements on mathematical benchmarks. 
Nevertheless, mathematical problem-solving remains a significant challenge, as current state-of-the-art LLMs lack universally applicable strategies for reliable external tool integration, necessitating development of more adaptive methodologies.

Retrieval-augmented generation (RAG) represents an alternative approach for incorporating external knowledge, retrieving relevant documents to enhance accuracy, reduce hallucinations in knowledge-intensive tasks, and improve result verifiability~\citep{gao2023retrieval, asai2023self}. Complex mathematical problems frequently require specialized or temporally current information not encoded in model parameters, establishing RAG as an essential complementary technique. Basic RAG implementations retrieve documents based on query embedding similarity, directly appending retrieved content to input prompts. Advanced RAG methodologies incorporate pre-retrieval operations (e.g., query reformulation) and post-retrieval processes (e.g., relevance re-ranking, content filtering) to optimize retrieved information quality and applicability~\citep{gao2023retrieval}. Through external knowledge integration, RAG facilitates access to relevant theorems, definitions, and established problem-solving strategies beyond model parametric knowledge.

Empirical evidence demonstrates RAG effectiveness in mathematical reasoning contexts. \citet{ragoproof} established that knowledge base retrieval effectively guides accurate proof generation. \citet{ragimprovemath} documented improvements in faithfulness and grounding, particularly for theorem-dependent problems. \citet{ragmathword} demonstrated that step-wise retrieved references enhance mathematical word problem performance through concrete fact anchoring of reasoning processes. Recent methodological advances~\citep{jin2025search, chen2025learning} have explored reinforcement learning approaches for training LLMs in dynamic information retrieval and integration. Active exploration and verification mechanisms in RAG-driven reasoning constitute promising research directions, as enhanced model capabilities for external knowledge identification, assessment, and integration hold significant potential for advancing mathematical problem-solving performance.
 
\section{Discussions}\label{sec:discussion}
The rapid advancement of mathematical reasoning in LLMs has revealed fundamental insights about the nature of reasoning capabilities, training methodologies, and inherent limitations. This section synthesizes key findings and implications across multiple dimensions of this evolving field.

\paragraph{Fundamental Nature of Chain-of-Thought Reasoning}
\label{subsec:fundamental}
CoT represents more than a prompting technique. It fundamentally enables models to learn and replicate structured procedures or algorithms rather than merely predicting outcomes. Evidence from diverse domains supports this interpretation: in navigation tasks, \citet{yang2022chain} and \citet{lehnert2024abetterplanningtransformers} demonstrated that CoT helps LLMs internalize planning and search algorithms through step-by-step decomposition. Similarly, the SYNAPSE framework~\citep{zheng2024synapse} applies CoT to web agent tasks through trajectory-as-exemplar prompting, enabling models to infer multi-step procedures from contextual demonstrations. This procedural learning capability positions CoT as a blueprint for reliable algorithmic problem-solving.

The expressive power of CoT, however, faces theoretical boundaries. Research by \citet{merrill2024the} establishes that the number of intermediate steps critically determines a transformer's computational capability: logarithmic steps provide modest enhancements, linear steps enable recognition of all regular languages when combined with projected pre-norm, while polynomial-time problems require computationally expensive polynomial steps. Alternative architectures, such as looped transformers~\citep{yang2024loopedtransformersbetterlearning}, offer potential solutions by incorporating iterative characteristics while maintaining efficiency with significantly reduced parameter counts.

\paragraph{Training Paradigm Trade-offs and Challenges}
\label{subsec:tradeoff}
The evolution from supervised fine-tuning to reinforcement learning reveals critical trade-offs in model development. While SFT significantly enhances instruction-following capabilities through alignment with human-annotated data, it simultaneously induces a diversity-alignment dilemma. Extensive SFT reduces generation diversity~\citep{li2025preserving, wang2planning}, potentially causing mode collapse where models produce consistently similar outputs. This diversity reduction proves particularly detrimental for downstream reinforcement learning applications requiring broad exploration spaces~\citep{zeng2025simplerl}. Recent evidence suggests that bypassing SFT in favor of direct RL optimization preserves exploratory capabilities, yielding superior performance in reasoning tasks~\citep{guo2025deepseek, zeng2025simplerl}.

Reward modeling presents another fundamental challenge. While Process Reward Models demonstrate superiority over Outcome Reward Models in reasoning tasks, obtaining high-quality reward signals remains prohibitively expensive. Traditional approaches rely on costly human-annotated datasets like PRM800k~\citep{lightman2023let}. Recent automated annotation methods using Monte Carlo sampling and MCTS~\citep{wang2024math} or process preference reward models~\citep{guan2025rstar} reduce annotation costs but remain vulnerable to reward hacking and struggle with generating precise reward scores~\citep{guo2025deepseek}.

\paragraph{Boundaries of Reasoning Improvement}
\label{subsec:boundaries}
A notable insight emerging from recent studies concerns the fundamental ceiling of reasoning improvements through RL. The insight lies in how different RL training paradigms affect model performance, as measured by the Pass@k metric.  For example, vanilla RL tends to improve Pass@1 performance while leaving Pass@k largely unchanged~\citep{shao2024deepseekmath}. In contrast, long CoT RL leads to gains in Pass@k metrics, reflecting enhanced quality and diversity across the top responses~\citep{zeng2025simplerl}. This improvement is attributed to longer generated sequences, which allow for the exploration of a broader set of reasoning pathways.

Meanwhile, multiple investigations~\citep{gandhi2025cognitive, edward2025demy, zeng2025simplerl, zhao2025echo, yue2025does} consistently report that such gains are primarily a result of activating and leveraging reasoning capabilities already present in the base models, rather than instilling fundamentally new abilities. For instance, \citet{gandhi2025cognitive} and \citet{edward2025demy} show that models such as Qwen inherently possess strong verification and backtracking skills, which RL simply helps to reveal. Similarly, \citet{yue2025does} empirically demonstrates that the upper bounds of sample performance during RL optimization are dictated by the capabilities of the base model, as reflected in Pass@k scores. These findings suggest that RL serves more as a mechanism for unlocking latent knowledge than for expanding a model’s reasoning capacity.

Nonetheless, it is questionable whether Pass@k (and the choice of $k$ in particular) is an appropriate and comprehensive measure of both base and fine-tuned model capabilities. If the metric does not fully capture the nuances of reasoning ability, then conclusions drawn solely from Pass@k may be limited in their validity.

\paragraph{Integration with Structural Search}
\label{subsec:integration}
The integration of structural search methods with LLMs represents a paradigm shift from single-path generation to systematic exploration with strategic backtracking capabilities. Structural search provides a solid approach for reasoning. This hybrid approach, combining classical algorithms with modern language models, demonstrates significant potential for tackling complex mathematical challenges~\citep{tot,guan2025rstar}. However, the fundamental constraint imposed by base model capabilities suggests that future advances may require innovations in pre-training methodologies or architectural designs rather than solely relying on fine-tuning optimization techniques.

These findings collectively indicate that while current methods effectively unlock and organize existing model capabilities, transcending these boundaries will likely require fundamental advances in how models acquire and represent mathematical knowledge during initial training phases. The field stands at a critical juncture where understanding these limitations can guide more targeted research efforts toward genuinely expanding reasoning capabilities rather than merely optimizing their expression.

\section{Potential Directions in Mathematical Reasoning for LLMs}
\label{sec:challenge}
Mathematical reasoning constitutes a fundamental component in the progression of Large Language Models toward Artificial General Intelligence. Despite substantial advances, significant challenges persist in computational approaches to mathematical problem-solving. This section examines three critical research directions: (1) extending LLM performance boundaries (Sections \ref{subsec:internalknwoledge} and \ref{subsec:externalknwoledge}), (2) enhancing reasoning efficiency (\ref{subsec:efficienttraining}), and (3) enabling cross-domain reasoning generalization (Section \ref{subsec:reasoning generalization}). Each direction presents distinct challenges requiring systematic investigation of current limitations and potential solutions.

\paragraph{Enhancing Reasoning through RL Exploration}
\label{subsec:internalknwoledge}
The emergence of DeepSeek-R1 demonstrates the transformative potential of long CoT reinforcement learning in advancing mathematical reasoning capabilities, particularly for multi-step problems including competition-level mathematics and complex logical derivations. While RL optimization aims to maximize base model capabilities, current implementations remain substantially below theoretical performance ceilings. Empirical evidence indicates that policies exhibit tendency toward reinforcing common reasoning patterns with limited exploration diversity~\citep{xiong2025minimalist}, resulting in convergence to local optima. The fundamental challenge involves efficient exploration of diverse reasoning pathways within constrained sampling budgets.

A promising research direction involves \textbf{leveraging compact representation spaces for exploration}. This approach constructs latent actions within low-dimensional spaces to guide token generation more efficiently. For instance, BWArea~\citep{jia2024bwarea} and CoLA~\citep{jia2025controlling} utilize future information to infer latent actions, enabling operation within compact action spaces that facilitate more comprehensive information gathering compared to conventional token-level optimization. This design paradigm offers potential for achieving more efficient exploration while maintaining computational tractability.

\paragraph{Knowledge-Augmented Reasoning}
\label{subsec:externalknwoledge}
Complex mathematical problems extending beyond model parametric knowledge—including open problems and mathematical conjectures—necessitate integration of external knowledge sources. The incorporation of external tools and retrieval-augmented generation represents a critical direction for expanding problem-solving capabilities~\citep{shen2024llm, gao2023retrieval}. External tools encompassing calculators, code interpreters, and geometric visualizers provide essential augmentation layers. The key capability requirement involves determining optimal timing and methodology for external resource invocation, requiring models to accurately assess their limitations and strategically leverage external assistance.

A primary research direction involves \textbf{training LLMs for effective external module interaction}. Recent investigations have explored optimization of interaction strategies through long CoT RL to minimize erroneous or unproductive external calls during reasoning processes~\citep{jin2025search, chen2025learning, li2025torl}. However, scaling challenges emerge as the number of available modules increases, each presenting unique capabilities and constraints. The \textbf{learnware paradigm}~\citep{zhou2024learnware} offers a structured solution through formalized registration, retrieval, and reuse mechanisms for specialized modules. By encapsulating external interactions into modular capabilities, this approach provides scalable pathways for enhancing reasoning performance across diverse task domains.

\paragraph{Optimality of Reasoning Paths}
\label{subsec:efficienttraining}
Long CoT generation frequently produces sub-optimal reasoning trajectories due to optimization processes prioritizing final answer correctness over path efficiency~\citep{sui2025stop}. Outcome-based reward functions evaluate terminal outputs while neglecting intermediate reasoning quality, treating responses as correct regardless of logical errors within reasoning chains. This approach fails to capture the importance of intermediate steps for achieving computational efficiency and maintaining logical consistency. Consequently, LLMs often generate unnecessarily lengthy or convoluted inference paths despite reaching correct conclusions~\citep{wang2025thoughts}.

Addressing this limitation requires \textbf{verification-aware optimization strategies}. Formal language integration with built-in proof systems (e.g., Lean) enables models to perform self-verification of intermediate steps, ensuring logical consistency while improving reasoning efficiency. Additionally, external tool integration for intermediate step validation provides mechanisms for solution correctness verification. However, effective tool utilization within long CoT frameworks for accuracy enhancement remains an active area of investigation requiring further theoretical and empirical development.

\paragraph{Reasoning Generalization to Open Domains}
\label{subsec:reasoning generalization}
While LLMs demonstrate cross-domain generalization within mathematics-such as solving AIME24 problems following MATH dataset training-robust generalization to open-ended domains presents fundamental challenges. Empirical investigations reveal that LLMs internalize abstract, compositional reasoning structures termed meta-skills, enabling flexible reuse of high-level cognitive patterns across contexts~\citep{guo2024learningpatternmatchingassaying, didolkar2024metacognitivecapabilitiesllmsexploration}. However, effective triggering and utilization of these meta-skills in complex, open-ended scenarios remains unresolved.

CoT prompting facilitates meta-level problem-solving pattern learning, providing limited task adaptation flexibility. This approach proves insufficient for broader generalization due to models' inability to internalize foundational reasoning principles including systematic abstraction and ideological thinking. Consequently, models rely on surface-level statistical correlations rather than adaptive reasoning for dynamic tasks such as scientific hypothesis generation. Domain-specific training becomes essential for addressing open-domain applications, with the primary challenge residing in generalizable reward function design. Unlike mathematics with clear binary feedback, complex real-world scenarios demand nuanced reward mechanisms for effective higher-order reasoning skill development.

Two promising research directions emerge: (1) \textbf{Reward model scaling}, inspired by pretraining scaling laws, involves expanding reward model training scales through learning from diverse pretraining datasets; (2) \textbf{Generative reward models} leverage generative architectures to construct reward functions, applying long CoT RL for progressive cultivation of complex, generalizable critic capabilities. These approaches offer potential pathways for bridging the gap between mathematical reasoning proficiency and general-purpose problem-solving capabilities.

\section{Conclusion}
This survey provides a comprehensive examination of mathematical reasoning in Large Language Models, organizing recent advances within a unified framework that distinguishes between comprehension (problem understanding) and generation (solution synthesis) capabilities. We  analyze the progression from training-free prompting techniques to sophisticated fine-tuning and inference-time scaling methods, revealing key insights about the current state and future directions of the field.

We identify three critical challenges defining the current frontier: (1) efficient exploration within constrained sampling budgets during RL optimization, (2) sub-optimal reasoning trajectories despite correct answers, and (3) limited generalization from mathematical to open-domain problem-solving. Promising research directions include compact representation spaces for exploration, verification-aware optimization, and scalable reward modeling.

As mathematical reasoning remains a critical benchmark for evaluating genuine understanding in the pursuit of AGI, this survey aims to serve as both a comprehensive reference for researchers and an accessible entry point for newcomers. We hope this work accelerates progress by providing a clear understanding of achievements, limitations, and the path forward in this vital area of AI research.

\bibliography{references}

\appendix
\newpage
\section{Mathematical Reasoning Evaluation Benchmarks}
\label{sec:evaluate}

In this section, we introduce benchmarks that can be used to evaluate a model's capabilities relevant to mathematical reasoning. An essential goal in evaluating mathematical reasoning models is to assess whether they exhibit capabilities comparable to, or exceeding, those of humans. To enable targeted evaluation, we propose categorizing datasets based on corresponding human levels, which allows us to assess different aspects of mathematical reasoning ability. These benchmarks are categorized by their difficulty levels, including elementary level, middle/high school level,  university level and competition level.

\subsection{Elementary Level}
\label{subsubsec:evaluate-math-elementary}
In the elementary level, MAWPS \citep{koncel2016mawps} is an online repository of 3,320 math word problems designed to evaluate models for solving such problems, covering topics such as basic arithmetic, algebra, and proportional reasoning. ASDiv-A \citep{miao2021diverse} consists mainly of elementary-level English math word problems, which encompass a range of topics such as arithmetic, algebra, and basic geometry concepts. Based on ASDiv-A, SVAMP \citep{patel2021nlp} is also composed of elementary-level math word problems, typically taught in grade four or lower. 
Math23K \citep{wang2017deep} is a collection of 23,161 Chinese elementary math word problems designed to evaluate the mathematical reasoning and problem-solving capabilities of models, particularly their ability to translate natural language into mathematical expressions and equations. 
MMLU \citep{hendrycks2020measuring}  is a dataset comprising 231,400 entries, with STEM tasks covering subjects such as mathematics, physics, computer science, and more.  Chinese Elementary School Math Word Problems (CMATH) dataset \citep{CMATH}, 
comprising 1.7k elementary school-level math word problems with detailed annotations, source from actual Chinese workbooks and exams.
\subsection{Middle/High School Level}
\label{subsubsec:evaluate-math-mh}

In the middle/high school level, GSM8K \citep{cobbe2021training} is a collection of 8.5K high-quality linguistically diverse grade school math word problems, designed to evaluate the mathematical reasoning capabilities of models, particularly their ability to solve multi-step problems that require understanding and manipulation of numbers, operations, and basic mathematical concepts. This benchmark assesses models' proficiency in areas such as basic arithmetic and algebra. Based on GSM8K dataset, GSM-HARD \citep{gao2023pal}, GSM-IC \citep{shi2023large}, GSM8K\_robust \citep{chern2023generative}, GSM-PLUS \citep{li2024gsm} add synthetic perturbation for further evaluation on models' capabilities of robustness, especially on arithmetic variation and critical thinking. By translating 250 math problems from GSM8K into different languages, researchers obtained Multilingual Grade School Math(MGSM)\citep{hendrycks2measuring}, with MGSM-zh commonly used as a Chinese dataset. 

\subsection{University Level}
\label{subsubsec:evaluate-math-challenge}
In university level benchmark, 
MathQA \citep{amini2019mathqa} is gathered by using a new representation language to annotate over the AQuA-RAT \citep{ling2017program} dataset with fully-specified operational programs. MathQA-Python \citep{austin2021program} is a Python adaptation of the MathQA benchmark, designed to evaluate LLMs' ability to synthesize code that solves mathematically complex problems described in natural language. OCW \citep{DBLP:journals/corr/abs-2206-14858} is a dataset of over 200
 undergraduate-level questions in science and mathematics from MIT’s OpenCourseWare to measure the model’s quantitative reasoning abilities in a CoT context. The SAT \citep{azerbayevllemma}dataset contains 32 math questions that do not contain figures from the May 2023 College Board SAT examination.

Additionally, there are benchmarks focusing on theorem proving and formal mathematics. For example, TheoremQA \citep{chen2023theoremqa} is a theorem-driven question-answering benchmark designed to assess LLMs' abilities to apply domain-specific theorems in solving complex problems. The LeanDojo \citep{yang2024leandojo} benchmark evaluates the theorem-proving capabilities of models by assessing their ability to generalize to novel theorems, which are not reliant on previously encountered premises. The benchmark includes 98,734 theorems/proofs extracted from mathlib, covering topics primarily at the university level. The miniF2F \citep{zheng2021minif2f} benchmark consists of 488 formal mathematics problems of Olympiad-level difficulty, offering a challenging and comprehensive evaluation of neural theorem-proving capabilities. 

\subsection{Competition Level}
\label{subsec:competition}
MATH \citep{hendrycks2measuring} consists of 12,500 competition-level mathematics problems spanning a range of topics, including algebra, geometry, calculus, and number theory. miniF2F \citep{zheng2021minif2f} is also an Olympiad-level benchmark. AIME24 (American Invitational Mathematics Examination 2024)~\citep{aime2024} consists of 30 problems from the 2024 American Mathematics Invitational, designed to evaluate models' performance on complex mathematical problems. OlympiadBench~\citep{he2024olympiadbench} is a bilingual, multimodal benchmark designed for Olympiad-level science competitions, featuring 8,952 problems from mathematics and physics Olympiads, including the Chinese college entrance exam.  MATH-Vision \citep{mathvision} provides a comprehensive and diverse set of challenges for evaluating  Large Multimodal Models'(LMM) mathematical reasoning abilities in mathematical reasoning within visual contexts.

\begin{table}
    \centering
    \small
    \setlength{\tabcolsep}{0.5mm}
    \setlength{\extrarowheight}{2pt}
    \caption{Mathematical reasoning benchmarks. \textbf{Synth.} indicates synthetic data. See tags below.}
    
    \smallskip
    \Tag{cyan}{M} = MWP, \Tag{red}{A} = Arithmetic, \Tag{purple}{G} = Geometry, \Tag{brown}{T} = Theorem Proving

    \begin{tabular}{l|llllll}
        \midrule \rowcolor{gray!20}
        \textbf{Level} &\textbf{Dataset} & \textbf{Domain}  & \textbf{Synth.} & \textbf{Language} & \textbf{Size} &\textbf{Release Time} \\
        \midrule
        \multirow{5}{*}{Elementary School} &
        MAWPS\citep{koncel2016mawps} & \Tag{cyan}{M}  & \XSolidBrush &En& 3,320 & Jun-2016\\
        &ASDiv-A\citep{miao2021diverse} & \Tag{cyan}{M} &  \XSolidBrush &En& 2,305 &Jul-2020\\
        & SVAMP\citep{patel2021nlp} & \Tag{cyan}{M}  & \ding{51} &En&1,000 &Apr-2021\\
        &Math23K\citep{wang2017deep} & \Tag{cyan}{M} & \XSolidBrush &Zh& 23,161 &Sept-2017\\

        &CMATH\citep{CMATH}&\Tag{cyan}{M}&\XSolidBrush&Zh&1,689&Jun-2023\\

        \midrule
        \multirow{4}{*}{Middle/High School} &
        GSM8K\citep{cobbe2021training} & \Tag{cyan}{M}  & \XSolidBrush &En& 8.5k & Oct-2021\\
        & GSM8K-PLUS\citep{li2024gsm} & \Tag{cyan}{M} &  \ding{51} &En& 10,552 &Feb-2024 \\
        &LILA\citep{mishra2022lila} & \Tag{cyan}{M}/\Tag{red}{A}/\Tag{purple}{G}/\Tag{brown}{T} &  \XSolidBrush &En&  133,815&Oct-2022\\
        
        &MGSM\citep{DBLP:journals/corr/abs-2210-03057} & \Tag{cyan}{M}  & \XSolidBrush &Multi& 250 & Oct-2022\\

        \midrule
        \multirow{7}{*}{University} & 
        AQuA\citep{ling2017program} & \Tag{cyan}{M} & \XSolidBrush &En& 100,000 &May-2017\\
        & MathQA\citep{amini2019mathqa} & \Tag{cyan}{M}  & \XSolidBrush &En& 37,297 &May-2019\\
        & SAT\citep{azerbayevllemma} & \Tag{cyan}{M}/\Tag{red}{A} &  \XSolidBrush &En& 32 &Oct-2023\\
        &OCW\citep{DBLP:journals/corr/abs-2206-14858}&\Tag{cyan}{M}&\ding{51}&En& 200&Jun-2022\\
        & TheoremQA\citep{chen2023theoremqa} & \Tag{purple}{G}/\Tag{brown}{T}  & \ding{51} &En& 800 &May-2023\\

        &LeanDojo\citep{yang2024leandojo} & \Tag{brown}{T} & \XSolidBrush &En& 98,734 &Jun-2023\\
        &miniF2F\citep{zheng2021minif2f} & \Tag{brown}{T} & \ding{51} &En& 488&Aug-2021\\
        \midrule
        \multirow{4}{*}{Competition} & 
        MATH\citep{hendrycks2measuring} & \Tag{cyan}{M}/\Tag{red}{A}/\Tag{purple}{G}/\Tag{brown}{T} & \XSolidBrush &En& 12,500 &Mar-2021\\
        &AIME24\citep{aime2024} & \Tag{cyan}{M}/\Tag{red}{A}/\Tag{purple}{G}/\Tag{brown}{T} & \XSolidBrush &En& 30 &Feb-2024\\
        & Olympicbench\citep{he2024olympiadbench} &  \Tag{cyan}{M}/\Tag{red}{A}/\Tag{purple}{G}/\Tag{brown}{T} & \XSolidBrush &En& 8.48k &Feb-2024\\
        &MATH-Vision\citep{mathvision}&\Tag{cyan}{M}/\Tag{red}{A}/\Tag{purple}{G}&\XSolidBrush&En&3,040&Feb-2024\\
        \midrule
        \multirow{1}{*}{Mixed}&MMLU-STEM\citep{hendrycks2020measuring}&\Tag{cyan}{M}/\Tag{red}{A}&\ding{51}&En&3,153&Sep-2020\\
    \bottomrule
    \end{tabular}
    \vspace{3mm}
    \label{mathbenchmark}
\end{table}

\end{document}